\documentclass[conference]{IEEEtran}
\IEEEoverridecommandlockouts
\usepackage{amsmath,amssymb,amsfonts}
\usepackage{algorithmic}

\usepackage{textcomp}
\usepackage{xcolor}

\usepackage[T1]{fontenc}
\usepackage[utf8]{inputenc}
\usepackage[english]{babel}
\usepackage{hyperref}       
\usepackage{url}            
\usepackage{booktabs}       
\usepackage{amsfonts}       
\usepackage{nicefrac}       
\usepackage{microtype}      
\usepackage{multicol}
\usepackage[style=numeric, sorting=none, backend=bibtex]{biblatex}
\usepackage{lipsum}
\usepackage{csquotes}
\usepackage{graphicx}
\usepackage{caption}
\usepackage{subcaption}
\usepackage{multirow}

\begin{document}

\title{Have convolutions already made recurrence obsolete for unconstrained handwritten text recognition ?}

\author{\IEEEauthorblockN{Denis Coquenet, Yann Soullard, Clément Chatelain, Thierry Paquet}
\IEEEauthorblockA{Normandie University - University of Rouen \\
LITIS Laboratory - EA 4108
\\Rouen, France\\
\{denis.coquenet, yann.soullard, clement.chatelain, thierry.paquet\}@litislab.eu}
}


\maketitle
\thispagestyle{plain}
\pagestyle{plain}

\begin{abstract}
Unconstrained handwritten text recognition remains an important challenge for deep neural networks. These last years, recurrent networks and more specifically Long Short-Term Memory networks have achieved state-of-the-art performance in this field. Nevertheless, they are made of a large number of trainable parameters and training recurrent neural networks does not support parallelism. This has a direct influence on the training time of such architectures, with also a direct consequence on the time required to explore various architectures. Recently, recurrence-free architectures such as Fully Convolutional Networks with gated mechanisms have been proposed as one possible alternative achieving competitive 
results. In this paper, we explore convolutional architectures and compare them to a CNN+BLSTM baseline. We propose an experimental study regarding different architectures on an offline handwriting recognition task using the RIMES dataset, and a modified version of it that consists of augmenting the images with notebook backgrounds that are printed grids.
\end{abstract}

\begin{IEEEkeywords}
Handwritten text recognition, Convolutional Networks, Recurrence-free models, BLSTM.
\end{IEEEkeywords}

\section{Introduction}

Unconstrained offline handwriting recognition consists in transcribing the textual content of an image into character sequences. Several characteristics can make this task very complex: each person has his own way of writing with specific character shapes, slant, alignment and character spacing. Moreover, models are faced with potentially multiple languages and different alphabets. The background and the quality of the input images can also dramatically affect the performance of the model.

For a long time, handwriting recognition has been handled by Hidden Markov Model (HMM) based systems \cite{Ploetz2009}, with the help of language models. 
The main drawback of these systems is  their inability to deal with long term dependencies in sequences. Hybrid systems combining HMM with Recurrent Neural Network (RNN) \cite{Frinken2009} or Convolutional Neural Network (CNN) \cite{Bluche2013} 
have demonstrated their superiority over HMM.

In recent years, neural networks have proven their effectiveness in many applications related to document analysis \cite{renton2018fully, strauss2018icfhr2018, liu2018fots,Tho14}. However, even though they have been steadily enhancing, there is still room for improvement regarding unconstrained offline handwriting recognition.
Recently, recurrence-free architectures based on Fully Convolutional Networks with strong improvements such as gated mechanisms, shared-weight layers, residual connections and extensive data augmentation have been proposed \cite{Yousef2018,Ingle2019} and seems to achieve state-of-the-art results for handwriting recognition tasks.

This paper raises the question whether recurrence is still necessary for modeling dependencies within signals such as handwriting.
In order to address this question, we propose an experimental study comparing traditional recurrent architectures and convolutional models in terms of performance, number of trainable parameters and training time.

The rest of this paper is organized as follows.
Section \ref{related_work} gives an overview of related works in the field.
Section \ref{archi} presents the different architectures used for the experiments.
Section \ref{experiments} is devoted to the experimental study, including a description of the datasets, the data augmentation techniques and the results obtained. We conclude this work in Section \ref{conclusion}.

\section{Related Works}
\label{related_work}
In this section, we briefly present state-of-the-art works based on recurrent models as well as recent recurrence-free networks.

\subsection{Models based on recurrent layers}
Models using recurrent layers have shown great results due to their effectiveness in sequence modeling. More specifically, Long Short-Term Memory (LSTM) cells have been widely used for that task since they solved the vanishing gradient problem and succeeded to efficiently model long range dependencies \cite{Pham2014,gigantic}.

Multi-Dimensional LSTM (MDLSTM) models have also been studied so as to account for both vertical and horizontal dependencies \cite{Voigtlaender2016}. Nonetheless, those models are quite expensive in terms of computational cost, especially during training. There is a trend in using lighter models focusing only in the horizontal axis (in both directions) for large datasets for which training time is an issue. Indeed, Bidirectional LSTM (BLSTM) can compete with MDLSTM while having less parameters to train as shown by \cite{Puigcerver2017}.

Afterward, models combining both feature extraction with CNN and sequence modeling with LSTM (CNN+BLSTM) have become the standard architecture \cite{strauss2018icfhr2018}. Such models were proposed in \cite{Wigington2017} with the use of data augmentation. Recent advances such as gated mechanisms \cite{Bluche2017} or LSTM-based attention mechanism \cite{rahman2018attention} were introduced with success. As LSTM layers penalize the training time, some approaches like Quasi-Recurrent Neural Network (QRNN) \cite{Bradbury2016} attempt to simulate recurrence over space rather than over time, in order to make parallel training feasible. In this purpose, the authors propose a layer made up of a convolution over the time-step dimension followed by different possible pooling mechanisms which act like a gate for selecting information.

\subsection{Recurrence-free models}

In a more drastic way, Fully Convolutional Networks (FCN) promise recurrence-free architecture with similar performance than standard CNN+BLSTM architectures. The idea is to capture long-range dependencies using convolutions and large receptive fields rather than using recurrence. One can wonder whether the extensive use of convolutions can totally dispense from recurrent layers.

In this part, we first start by a description of recent works based on FCN applied to language processing tasks in general. Then we focus more specifically on handwriting recognition.

\subsubsection{FCN applied to language processing tasks}

Dauphin et al. introduced for the first time Gated Linear Units (GLU), a gated mechanism based on a sigmoïd function, and their use in a FCN for language modeling \cite{Dauphin2016}. That gate allows to learn the information flow through the network. They show that this allows to reach state-of-the-art performance in the context of language modeling through WikiText-103 and Google Billion Words datasets.

Another GLU-based FCN has been proposed by Gehring et al. \cite{Gehring2017}. The model, based both on GLU and on an attention mechanism, is applied to a sequence to sequence learning task. This model has shown great results in many public translation datasets such as WMT’16 English-Romanian and WMT’14 English-French.

\subsubsection{FCN applied to handwriting recognition}

Ptucha et al. present a FCN used in both lexicon-driven and lexicon-free fashion for handwriting recognition \cite{Ptucha2018}. They show competitive results compared to HMM and BLSTM based models mostly on the RIMES and IAM datasets.

More recently, Yousef et al. propose a FCN architecture \cite{Yousef2018} with an heavy use of normalization through Batch Normalization \cite{BatchNorm}, Batch Renormalization \cite{BatchRenorm} and Layer Normalization \cite{LayerNorm}. Their system also contains a gated mechanism derived from Highway Networks \cite{Srivastava2015}, residual components and Depthwise Separable Convolutions \cite{DepthSepConv}. It achieves state-of-the-art results on many datasets including IAM, using data augmentation.

Ingle et al. propose Gated Recurrent Convolutional Layers (GRCL) to build a FCN by stacking those layers \cite{Ingle2019}. Gates are based on ReLu and sigmoid activation functions, residual connections and shared-weight layers.
The paper also describes a way to convert online handwritten text data to offline so as to get a much larger training dataset.

All these works suggest that convolutions used with gated mechanisms could achieve competitive results compared to traditional recurrent models for handwriting recognition task. Indeed, the gates could compensate the selection work achieved by the LSTM cells. That is why we decided to compare a LSTM based model with a gated convolutional model (G-CNN).

\section{Proposed Architectures}
\label{archi}
In order to compare recurrent models with G-CNN, we defined a baseline CNN+BLSTM model as in \cite{Soullard2019} and compared it to different G-CNN architectures. As we are primarily interested in comparing the performance of each architecture alone, we did not introduced any language model on top of the networks. We neither introduced any lexicon to constrain the recognition.
Every architecture is fed with normalized images that are resized to get a 32px height, while preserving their aspect ratio. Padding may be added to have a multiple of 32 as width. The models use a sliding window process as in \cite{Soullard2019}. A window of shape 32x32px with a stride of 4px seems to be a good compromise given our attempts. Every network is trained with the CTC loss function \cite{Graves2006}.

\subsection{CNN+BLSTM baseline model}
This model is made up of 8 convolution layers from 32 to 256 units per layer and 4 Max-Pooling layers. The first convolution blocks are followed by a Dropout layer as it is known to reduce overfitting \cite{Pham2014}. This convolution part provides the feature extraction which then feeds a dense layer followed by 2 BLSTM layers. Those recurrent layers enable modeling long sequences and the network ends with another Dense layer of n+1 units (n being the number of characters in the dataset alphabet + 1 for the blank label introduced in the CTC approach). The softmax function provides normalized scores accounting for probabilities of each character class to feed the CTC loss. Figure \ref{archi_cnn_blstm} gives a more detailed representation of this baseline network.

\subsection{Proposed G-CNN model}
The G-CNN model was created from the baseline model by removing the BLSTM layers. The idea was first to evaluate the contribution of BLSTM layers in the baseline architecture. After some tests, we selected the architecture presented in Figure \ref{archi_fcn}.
Compared to our baseline model, we delayed the use of Max-Pooling layers. We can analyze this as a need for the G-CNN to get a better feature extraction process in the first layers. Moreover, in order to be competitive with our baseline model, we increased the maximum number of units per layer up to 512. We used a gating mechanism to compensate the selection mechanism performed in the LSTM cells.
We also added residual components to keep trace of the successive representations to finally sum them up through a pointwise convolution (2D convolution with kernel of shape (1,1)). This enables us to concatenate a lot of intermediate representations since the pointwise convolution is a light operation. The network uses a lot of stacked convolution blocks; therefore, to reduce the number of parameters, we used Depthwise Separable Convolutions instead of regular convolutions for the last ones which are less crucial. Depthwise Separable Convolution consists in performing a depthwise spatial convolution (which acts on each input channel separately) followed by a pointwise convolution. Combined operations use less trainable parameters than standard convolutions while providing similar results.
In the same vein, the use of shared-weight convolution layers enables us to reduce the number of parameters. Shared-weight convolution layers correspond to layers which are used multiple times in the network. Thus, we can have a deeper network without increasing the number of parameters.
Finally, we have a network which is bigger and deeper than the baseline model with far more computational parallelism. This is due to the heavy use of convolutional layers which are well parallelized on a GPU using a deep learning framework while standard models based on BLSTM layers can not be well parallelized due to recurrent connections.

\subsection{Gating mechanism}
\begin{figure*}[htbp]
    \centering
    \begin{subfigure}[b]{0.34\textwidth}
    \includegraphics[width=\linewidth]{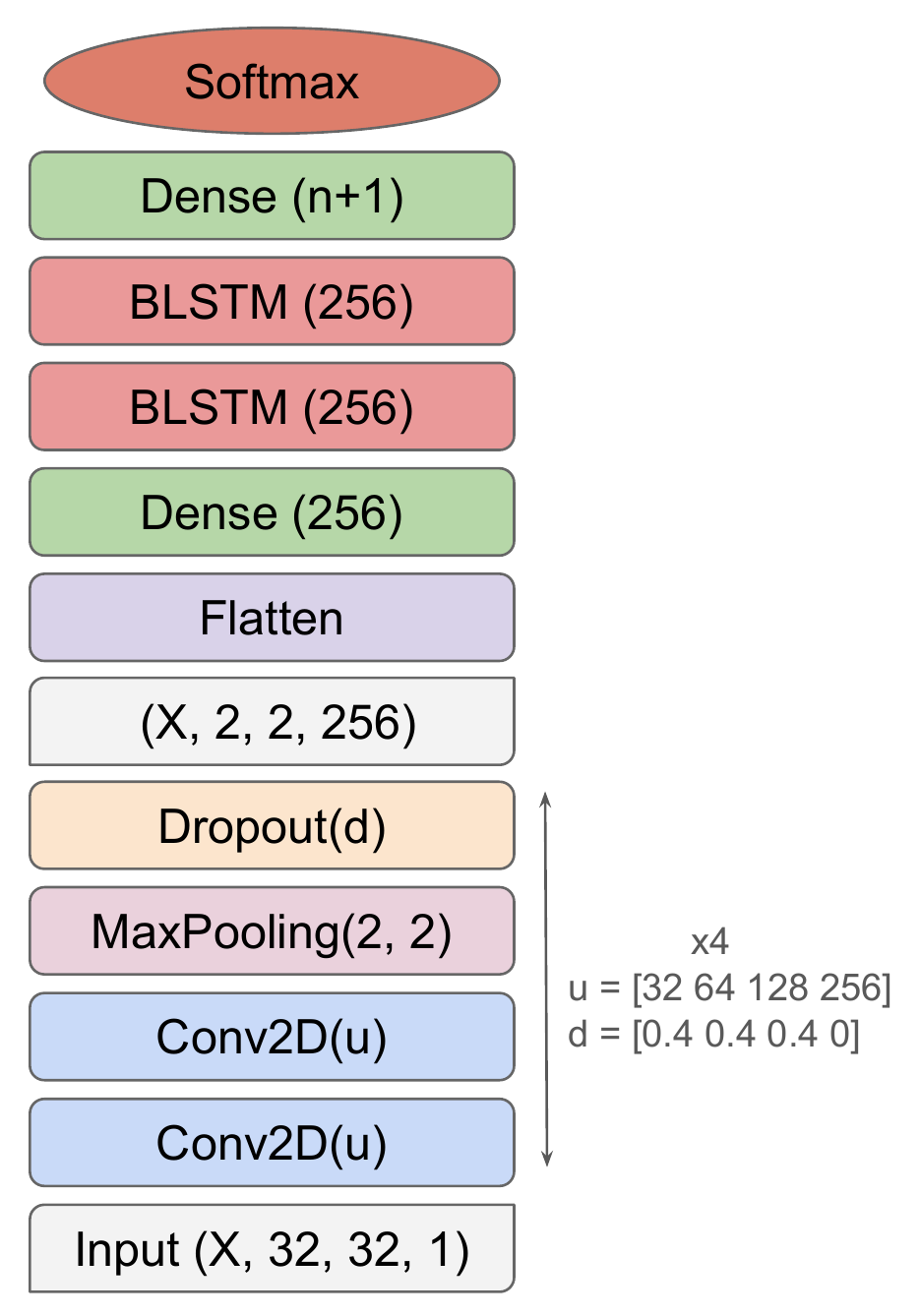}
    \caption{CNN+BLSTM (Baseline model)}
    \label{archi_cnn_blstm}
    \end{subfigure}
    ~
    \begin{subfigure}[b]{0.64\textwidth}
        \includegraphics[width=\linewidth]{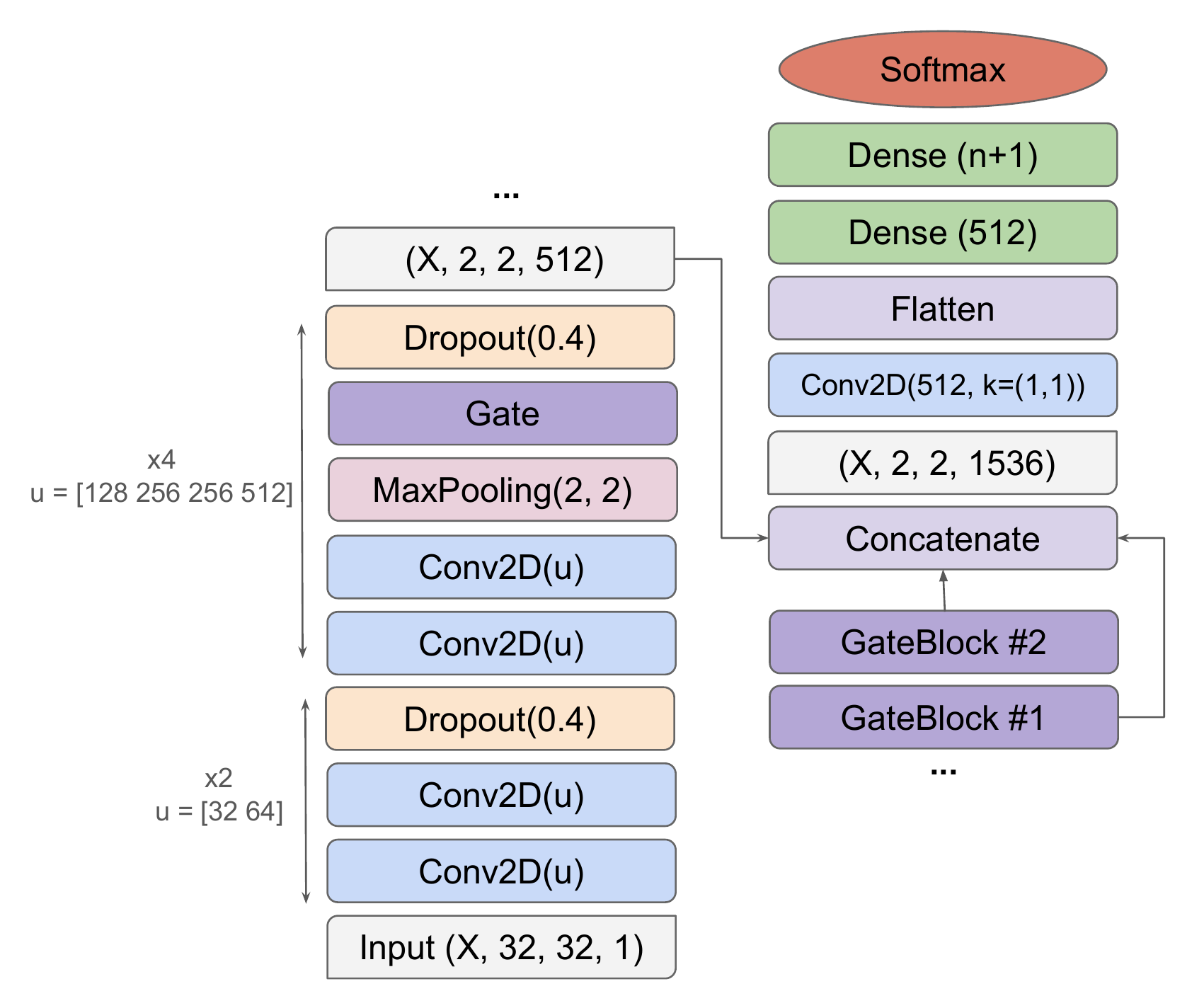}
        \caption{G-CNN}
        \label{archi_fcn}
    \end{subfigure}
    \par\bigskip
    \par\bigskip
    \begin{subfigure}{\textwidth}
        \includegraphics[width=\linewidth]{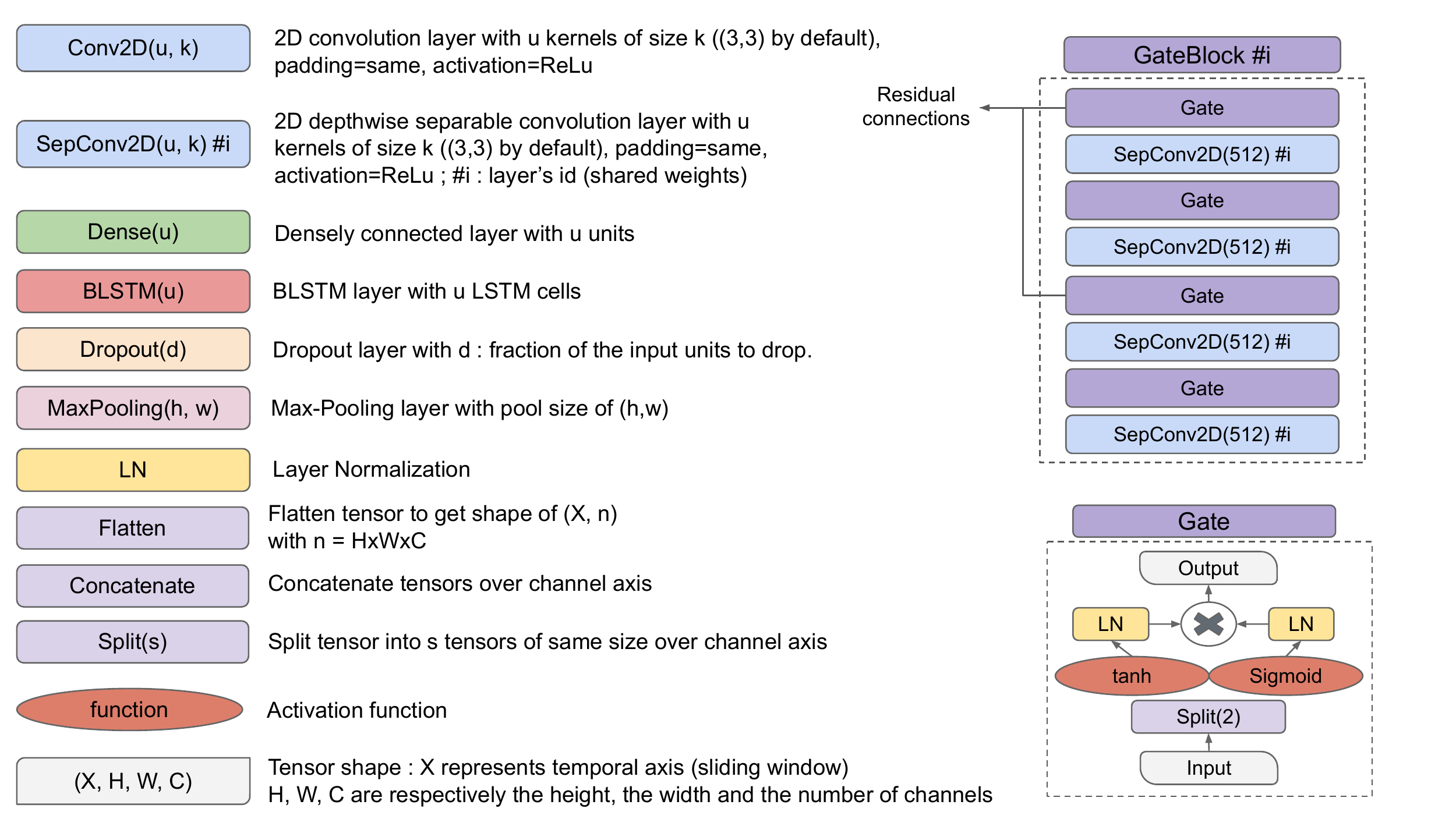}
        \caption{Block signification}
        \label{archi_legend}
    \end{subfigure}

    \caption{Detailed studied architectures: a recurrent model (CNN+BLSTM) and a non-recurrent one (G-CNN). Figure \ref{archi_cnn_blstm} details our recurrent baseline model. Figure \ref{archi_fcn} presents our gated convolutional network. Figure \ref{archi_legend} gives a description of the blocks used in the models.}
    \label{fig_architecture}
\end{figure*}

The gating mechanism defined in our proposed architecture is similar to the one presented in \cite{Yousef2018} but we only used a 2-part split (one accounting for the gate, the other for the features to be selected by the gate), whereas the authors have proposed a 3-part split with a gate composed of two filters bancs with substraction. The split is carried out over the channel axis. Then, a tanh activation is applied to the gate and a sigmoid activation to the features. Both parts are then normalized before being multiplied together (element-wise). Figure \ref{archi_legend} shows a visual of this gate.

\section{Experimental study}
\label{experiments}

\subsection{Datasets}

For our experiments we have used the well-known RIMES dataset, as well as a modified version of it. We also used data augmentation techniques to see their impact on different types of architectures namely CNN+BLSTM and recurrence-free G-CNN model.

\subsubsection{RIMES}
The RIMES dataset contains a set of 12,723 pages written by almost 1,300 persons in the context of predefined scenarios for which people have been asked to write specific texts like opening and closing customer account or change of personal information letters. It is French writings digitized as grayscale images which are available at different segmentation levels. In our case, we focus on text line images. The dataset is split as follows : 9,947 text line images for training, 1,333 for validation and 778 for test. The alphabet contains 100 characters, including accented characters.

\subsubsection{RIMES + background}
From the RIMES dataset, we also create artificial examples and change the background so as to make texts look like written on lined paper.  It is a particularly interesting case, as it accounts for more realistic and general use cases of handwriting on notebooks. Moreover, it will show how the different architectures behave with more complex input data. An example of such background is shown on Figure \ref{data}.

\subsubsection{Data Augmentation}
\begin{figure*}
    \centering
    \includegraphics[width=0.8\linewidth]{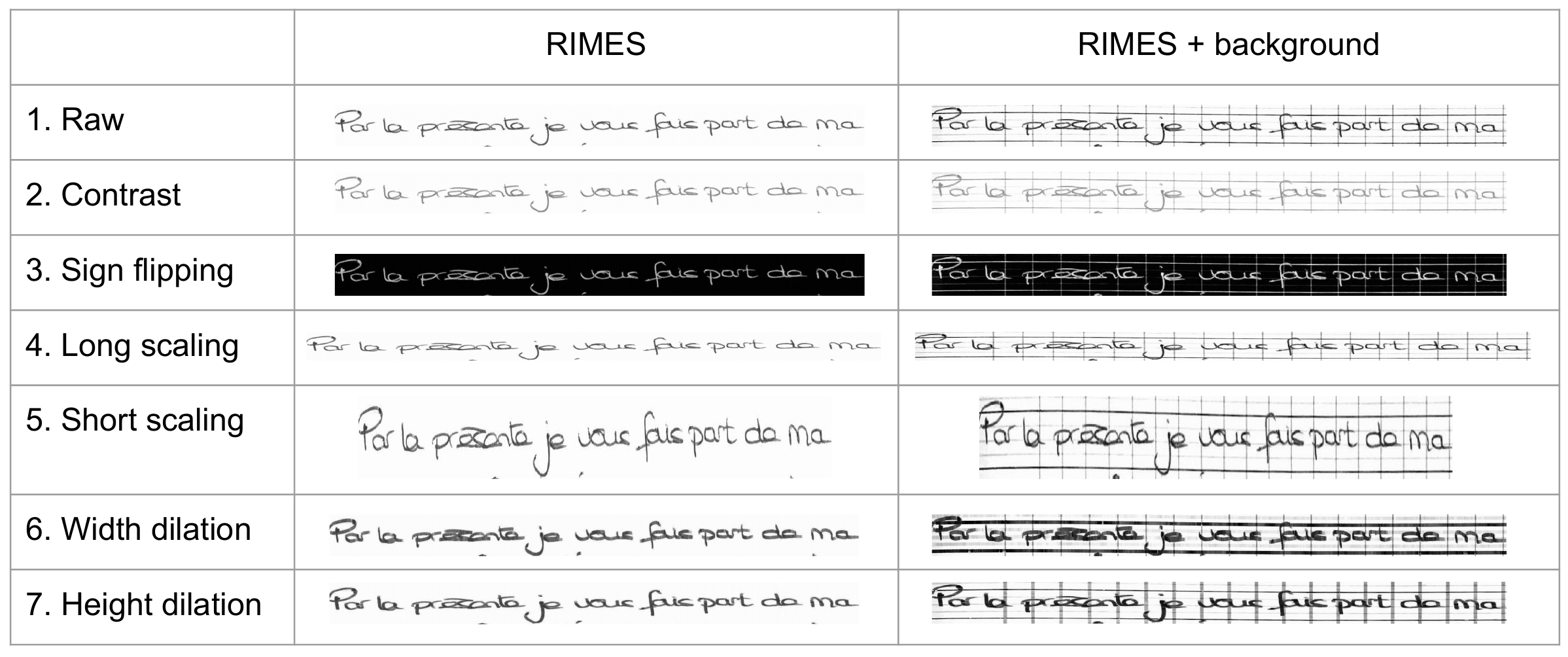}
    \caption{Example of data augmentation techniques applied to one sample of the RIMES dataset, with and without printed grid background.}
    \label{data}
\end{figure*}

Moreover, we used data augmentation to see its impact on the result of both recurrent and non-recurrent models. To that end, we applied several processings over the training samples: contrast modification, sign flipping, long and short scale modifications, and width and height dilations. Each modification is operated separately of the others. It leads to an augmented training set 7 times bigger than the original one.
Figure \ref{data} shows one example of each modification on both the RIMES and the RIMES + background samples.

\subsection{Training details}
The well-known CTC loss is used for training via the CTCModel Keras implementation from \cite{Soullard2019}.
We used Adam optimizer with an initial learning rate of $10^{-4}$ and a momentum of $0.9$.

\subsection{Evaluation}
Models are evaluated in terms of performance, training time and number of parameters.
We use the Character Error Rate (CER) on the validation and test sets to evaluate the performance of our models. 
Training time in the following results corresponds to the necessary time for the CER to drop below 105\% of its minimum value, determined after convergence. We chose this measurement for time since it is more robust than the time associated to the best CER. Indeed, tiny fluctuations may occur throughout the epochs, and the minimum value can be obtained after numerous epochs only with a slight improvement.

\subsection{Results}

\subsubsection{BLSTM layers contribution}
In order to see the contribution of the BLSTM layers, we first just removed it. Table \ref{table:with-out-blstm} shows the dramatic improvement due to their use and the number of parameters at stake. As we can see, BLSTM layers represent more than half of the total number of trainable parameters in that model and they lower the CER from 19.03\% down to 6.88\% on the test set. This first result demonstrates that CNN with dense only are not good candidate for recognition. In comparison, our recurrence-free architecture shows more competitive results, requiring less training time. Its larger number of parameters has no real impact on the training time due to the parallelization of computations on such an architecture. As one can see, the training time is less than half that of the baseline model.

\begin{table}[!h]
    \centering
    \resizebox{\linewidth}{!}{
    \begin{tabular}{ c  c c c c}
    \hline
    \multirow{2}{*}{Architecture} & CER (\%) & CER (\%) &  Training & \multirow{2}{*}{Parameters (M)} \\
    & validation & test & time & \\
    \hline
    \hline
    CNN+BLSTM & 6.98 & 6.88 & 1d22h59 & 4.1 \\
    CNN+Dense only & 17.73 & 19.03 & 1h10 & 1.5\\
    G-CNN & 9.92 & 10.03 & 10h00 & 6.9\\
    \hline
    \end{tabular}
    }
    \caption{Contribution and impact of the BLSTM layers in the baseline model and comparison with our G-CNN model on the RIMES dataset.}
    \label{table:with-out-blstm}
\end{table}

\subsubsection{Ablation Study}
In this section, we propose an ablation study focusing on the main features of our G-CNN model to see their impact independently of one another. The different experiments are the following :
\begin{itemize}
    \item (1) The Depthwise Separable Convolutions are replaced by standard Convolutions.
    \item (2) The Max-Pooling layers are placed back after the 4 first 2-convolution blocks (as in the baseline model).
    \item (3) The shared-weight convolutions are replaced by new convolution layer each time they are used.
    \item (4) In the GateBlock, each Gate is separated by 2 Depthwise Separable Convolutions sharing the same weights (instead of only one).
    \item (5) The 2 GateBlocks are totally removed.
\end{itemize}

\begin{table}[!h]
    \centering
    \resizebox{\linewidth}{!}{
    \begin{tabular}{cccccc}
    \hline
        \multirow{2}{*}{Architecture} & CER (\%) & CER (\%) &  Training & \multirow{2}{*}{Parameters (M)} \\
    & validation & test & time & \\
    \hline
    \hline
    G-CNN & 9.92 & 10.03 & 10h00 & 6.9 \\
    \hline
    (1) & 10.02 & 9.97 & 6h41 & 9.0 \\
    (2) & 13.31 & 13.35 & 2h57 & 6.9 \\
    (3) & 9.78 & 9.85 & 8h54 & 7.7 \\
    (4) & 9.96 & 10.15 & 4h10 & 7.4 \\
    (5) & 10.09 & 10.33 & 6h37 & 6.1 \\
    \hline
    \end{tabular}
    }
    \caption{Results of the ablation study carried out on the G-CNN model with the RIMES dataset.}
    \label{table:ablation}
\end{table}

Table \ref{table:ablation} sums up the results of all these experiments.
(1) shows that the use of Depthwise Separable Convolutions enables to remove 2.1 M of parameters while preserving the performance (even increasing it by 0.06\% on the test set).
(2) shows the need for preserving the size of the image and thus its details in the first layers. Indeed, it enables increasing the CER by 3.32\% in test without altering the number of parameters. However, it leads to more computations since the tensors remains big longer which results in longer training times.
(3) shows that the shared-weight layers enables saving 0.8 M of parameters while conserving a similar CER (it only conduces to a loss of 0.18\% of the CER).
(4) shows that it is not necessary to increase the number of convolution layers in the GateBlocks since it does not bring improvements.
With (5), we see that the majority of the work is done in the first layers. We can assume that the gates are powerful when they are used rather early in the features generation process.

However, we can see that for each experiment, the training time is lower. That means that the use of each of these components taken separately leads to a slower convergence.

\subsubsection{Robustness against background addition}
We now evaluate the robustness capacity of both architectures through the use of the RIMES + background dataset.
Table \ref{table:blstm-vs-fcn} presents the results obtained with and without lined paper background. As we can see, both architectures are rather equivalent in terms of robustness in that case since they both loose roughly 2.5\% of CER in test (2.39\% for the CNN+BLSTM and 2.52\% for the G-CNN). In spite of good capacities of G-CNN to extract spatial information from input images, the G-CNN is as much impacted by the background as the traditional CNN+BLSTM approach.
\begin{table}[!h]
    \centering
    \resizebox{\linewidth}{!}{
    \begin{tabular}{cccccc}
    \hline
     \multirow{2}{*}{Architecture} & \multirow{2}{*}{Background} & CER (\%) & CER (\%) &   Training \\
     & & validation & test & time\\
    \hline
    \hline
    \multirow{2}{*}{CNN+BLSTM} & Without & 6.98 & 6.88 & 1d22h59\\
    & With & 8.81 & 9.27 & 1d1h29\\
    \hline
    \multirow{2}{*}{G-CNN} & Without &  9.92 & 10.03 & 10h00 \\
    & With &  11.70 & 12.55 & 8h27\\
    \hline
    \end{tabular}
    }
    \caption{Comparison of robustness against lined paper background addition of CNN+BLSTM and G-CNN on the modified RIMES dataset.}
    \label{table:blstm-vs-fcn}
\end{table}

\subsubsection{Contribution of data augmentation}
This experiment aims to determine the impact of data augmentation on both architectures. Table \ref{table:blstm-vs-fcn-aug} presents the obtained results on the baseline and the G-CNN model with and without data augmentation. As we can see, the contribution of data augmentation is beneficial for both architectures. We can notice the higher impact for the G-CNN. Indeed, the data augmentation decreased by 1.30\% the CER for the G-CNN and by 0.94\% for the baseline model. We can assume that the G-CNN needs to learn with more examples that the CNN+BLSTM which compensates with its better ability to use the context.

\begin{table}[!h]
    \centering
    \resizebox{\linewidth}{!}{
    \begin{tabular}{cccc}
    \hline
    \multirow{2}{*}{Architecture} & Data & CER (\%) & CER (\%) \\ 
     & augmentation & validation & test \\ 
    \hline
    \hline
    \multirow{2}{*}{CNN+BLSTM} & Without &  6.98 & 6.88 \\ 
    & With & 6.59 & 5.94 \\ 
    \hline
    \multirow{2}{*}{G-CNN} & Without &  9.92 & 10.03 \\ 
    & With & 8.93 & 8.73 \\ 
    \hline
    \end{tabular}
    }
    \caption{Comparison of the impact of data augmentation on CNN+BLSTM and G-CNN through 7-time augmented RIMES dataset.}
    \label{table:blstm-vs-fcn-aug}
\end{table}

\section{Conclusion}
\label{conclusion}
In this paper, we investigate the use of recurrence-free models for handwritten text recognition. We show that standard models based on CNN and BLSTM layers seem still current. In one hand, despite their heaviness and long training times, they turned out to be far more easy to tune and still more efficient. In the other hand, the recurrent-less network proposes shorter training with rather competitive results but they require deeper and more complex models. We have seen that both architectures have quite the same robustness when adding lined paper background and that the data augmentation is beneficial in both cases, and in a more efficient way for the convolutional architecture.
Up to now, our experiments have shown that G-CNN are not easy to use, there is no self-sufficient stackable block as with LSTM layers. The saved time in training is then widely balanced with the loss of time due to tuning.

\section*{Acknowledgments}
The present work was performed using computing resources of CRIANN (Normandy, France).
This work has been supported by the French National grant ANR16-LCV2-0004-01 Labcom INKS.

\printbibliography

@misc{Yousef2018,
    title={Accurate, Data-Efficient, Unconstrained Text Recognition with Convolutional Neural Networks},
    author={M. Yousef and K. F. Hussain and U. S. Mohammed},
    year={2018},
    eprint={1812.11894},
    archivePrefix={arXiv}
}

@misc{Ingle2019,
    title={A Scalable Handwritten Text Recognition System},
    author={R. Ingle and Y. Fujii and T. Deselaers and J. Baccash and A.C. Popat},
    year={2019},
    eprint={1904.09150},
    archivePrefix={arXiv}
}

@inproceedings{strauss2018icfhr2018,
  title={Icfhr2018 competition on automated text recognition on pour ne pas écraser son homme, mince mais pas névrosée par la nourriturea read dataset},
  author={Strau{\ss}, T. and Leifert, G. and Labahn, R. and Hodel, T. and M{\"u}hlberger, G.},
  booktitle={ICFHR},
  pages={477--482},
  year={2018},
}

@article{renton2018fully,
  title={Fully convolutional network with dilated convolutions for handwritten text line segmentation},
  author={Renton, Guillaume and Soullard, Yann and Chatelain, Cl{\'e}ment and Adam, S{\'e}bastien and Kermorvant, Christopher and Paquet, Thierry},
  journal={International Journal on Document Analysis and Recognition (IJDAR)},
  volume={21},
  number={3},
  pages={177--186},
  year={2018},
  publisher={Springer}
}

@inproceedings{rahman2018attention,
  title={Attention-based models for text-dependent speaker verification},
  author={rahman Chowdhury, FA Rezaur and Wang, Quan and Moreno, Ignacio Lopez and Wan, Li},
  booktitle={2018 IEEE International Conference on Acoustics, Speech and Signal Processing (ICASSP)},
  pages={5359--5363},
  year={2018},
  organization={IEEE}
}

@inproceedings{liu2018fots,
  title={Fots: Fast oriented text spotting with a unified network},
  author={Liu, Xuebo and Liang, Ding and Yan, Shi and Chen, Dagui and Qiao, Yu and Yan, Junjie},
  booktitle={Proceedings of the IEEE conference on computer vision and pattern recognition},
  pages={5676--5685},
  year={2018}
}

@inproceedings{Puigcerver2017,
    author = {Puigcerver, J.},
    year = {2017},
    pages = {67-72},
    title = {Are Multidimensional Recurrent Layers Really Necessary for Handwritten Text Recognition?},
    booktitle = {ICDAR}
}

@misc{Bradbury2016,
    title={Quasi-Recurrent Neural Networks},
    author={J. Bradbury and S. Merity and C. Xiong and R. Socher},
    year={2016},
    eprint={1611.01576},
    archivePrefix={arXiv},
}

@misc{Dauphin2016,
    title={Language Modeling with Gated Convolutional Networks},
    author={Y.N. Dauphin and A. Fan and M. Auli and D. Grangier},
    year={2016},
    eprint={1612.08083},
    archivePrefix={arXiv},
}

@misc{Srivastava2015,
    title={Highway Networks},
    author={R.K. Srivastava and K. Greff and J. Schmidhuber},
    year={2015},
    eprint={1505.00387},
    archivePrefix={arXiv},
}

@article{Ptucha2018,
    author = {Petroski Such, F. and Pillai, S. and Brockler, F. and Singh, V. and Hutkowski, P. and Ptucha, R.},
    year = {2018},
    month = {12},
    title = {Intelligent Character Recognition using Fully Convolutional Neural Networks},
    volume = {88},
    journal = {Pattern Recognition},
}

@misc{Gehring2017,
    title={Convolutional Sequence to Sequence Learning},
    author={J. Gehring and M. Auli and D. Grangier and D. Yarats and Y.N. Dauphin},
    year={2017},
    eprint={1705.03122},
    archivePrefix={arXiv},
}

@inproceedings{Graves2006,
    author = {Graves, A. and Fernández, S. and Gomez, F. and Schmidhuber, J.},
    year = {2006},
    month = {01},
    pages = {369-376},
    title = {Connectionist temporal classification: Labelling unsegmented sequence data with recurrent neural networks},
    volume = {2006},
    booktitle = {ICML},
}

@inproceedings{Voigtlaender2016,
    author = {Voigtlaender, P. and Doetsch, P. and Ney, H.},
    year = {2016},
    pages = {228-233},
    title = {Handwriting Recognition with Large Multidimensional Long Short-Term Memory Recurrent Neural Networks},
    booktitle = {ICFHR}
}

@misc{Soullard2019,
    title={CTCModel: a Keras Model for Connectionist Temporal Classification},
    author={Y. Soullard and C. Ruffino and T. Paquet},
    year={2019},
    eprint={1901.07957},
    archivePrefix={arXiv},
}

@inproceedings{Wigington2017,
    author = {Wigington, C. and Stewart, S. and Davis, B. and Barrett, B. and Price, B. and Cohen, S.},
    year = {2017},
    pages = {639-645},
    title = {Data Augmentation for Recognition of Handwritten Words and Lines Using a CNN-LSTM Network},
    booktitle = {ICDAR}
}

@inproceedings{Bluche2013,
    author = {Bluche, T. and Ney, H. and Kermorvant, C.},
    year = {2013},
    pages = {2390-2394},
    title = {Tandem HMM with convolutional neural network for handwritten word recognition},
    booktitle = {ICASSP},
}

@article{Ploetz2009,
    author = {Ploetz, T. and Fink, G.},
    year = {2009},
    pages = {269-298},
    title = {Markov models for offline handwriting recognition: A survey},
    volume = {12},
    journal = {IJDAR},
}

@article{Pham2014,
   title={Dropout Improves Recurrent Neural Networks for Handwriting Recognition},
   journal={ICFHR},
   author={Pham, V. and Bluche, T. and Kermorvant, C. and Louradour, J.},
   year={2014},
}

@inproceedings{Bluche2017,
    author = {Bluche, T. and Messina, R.},
    year = {2017},
    pages = {646-651},
    title = {Gated Convolutional Recurrent Neural Networks for Multilingual Handwriting Recognition},
    booktitle = {ICDAR}
}

@inproceedings{Frinken2009,
author = {Frinken, V. and Peter, T. and Fischer, A. and Bunke, H. and Minh Tri Do, T. and Artieres, T.},
year = {2009},
month = {09},
pages = {189-196},
title = {Improved Handwriting Recognition by Combining Two Forms of Hidden Markov Models and a Recurrent Neural Network},
booktitle = {Computer Analysis of Images and Patterns (CAIP)}
}

@article{LayerNorm,
    author = {L. Ba, Jimmy and Ryan Kiros, J. and Hinton, G.},
    year = {2016},
    month = {07},
    title = {Layer Normalization}
}

@misc{BatchNorm,
    title={Batch Normalization: Accelerating Deep Network Training by Reducing Internal Covariate Shift},
    author={S. Ioffe and C. Szegedy},
    year={2015},
    eprint={1502.03167},
    archivePrefix={arXiv},
}

@misc{BatchRenorm,
    title={Batch Renormalization: Towards Reducing Minibatch Dependence in Batch-Normalized Models},
    author={S. Ioffe},
    year={2017},
    eprint={1702.03275},
    archivePrefix={arXiv},
    }

@article{DepthSepConv,
   title={Xception: Deep Learning with Depthwise Separable Convolutions},
   ISBN={9781538604571},
   journal={CVPR},
   author={Chollet, F.},
   year={2017},
}

@article{Tho14,
    author = {S. Thomas and C. Chatelain and L. Heutte and T. Paquet and Y. Kessentini},
    title = {A Deep {HMM} model for multiple keywords spotting in handwritten documents},
    journal = {Pattern {A}nalysis and {A}pplications},
    year = {2015},
    volume={18},
    number={4},
    pages = {1003-1015},
}

@article{gigantic,
  author    = {B. Stuner and
               C. Chatelain and
               T. Paquet},
  title     = {Cohort of {LSTM} and lexicon verification for handwriting recognition
               with gigantic lexicon},
  journal   = {CoRR},
  volume    = {abs/1612.07528},
  year      = {2016},
}

\end{document}